\documentclass[sigconf]{acmart}
\usepackage{bbm}
\usepackage{xspace}
\usepackage{tcolorbox}
\usepackage{xcolor}
\usepackage{bm} 
\usepackage{multirow}
\usepackage{array}
\usepackage{booktabs}
\usepackage{amsfonts}
\usepackage{amsmath}
\usepackage{amsthm}
\usepackage{subfigure}
\usepackage{listings}
\usepackage{threeparttable}
\usepackage{algorithm}
\usepackage{algpseudocode}

\usepackage{diagbox}
\usepackage{pifont}
\usepackage{makecell}
\usepackage{graphicx}
\usepackage{threeparttable}
\usepackage{tcolorbox}
\newcommand{\ours}[0]{Hetero$^2$Net\xspace}

\newcommand{\sota}[1]{\textbf{#1}}

\definecolor{blue}{rgb}{0.047, 0.365, 0.647}
\definecolor{c1}{rgb}{0.88, 0.92, 96}
\definecolor{c2}{rgb}{0.65, 0.858, 0.922}
\definecolor{c3}{rgb}{0.24, 0.465, 1}
\definecolor{c4}{rgb}{0.16, 0.43, 0.76}
\definecolor{c5}{rgb}{0., 0.3, 0.76}
\newcommand{\nosection}[1]{\vspace{2pt}\noindent\textbf{#1.}}

\AtBeginDocument{%
  }

\setcopyright{acmcopyright}
\copyrightyear{2018}
\acmYear{2018}
\acmDOI{XXXXXXX.XXXXXXX}

\acmConference[Conference acronym 'XX]{Make sure to enter the correct
  conference title from your rights confirmation emai}{June 03--05,
  2018}{Woodstock, NY}
\acmPrice{15.00}
\acmISBN{978-1-4503-XXXX-X/18/06}

\begin{document}

\title{\ours: Heterophily-aware Representation Learning on Heterogeneous Graphs}


\author{Jintang Li}
\authornote{Both authors contributed equally to this research.}
\affiliation{\institution{Sun Yat-sen University}
    \country{}}
\email{lijt55@mail2.sysu.edu.cn}

\author{Zheng Wei}
\authornotemark[1]
\affiliation{\institution{Ant Group}
    \country{}}
\email{frank.wz@antgroup.com}

\author{Jiawang Dan}
\affiliation{\institution{Ant Group}
    \country{}}
\email{yancong.djw@antgroup.com}

\author{Jing Zhou}
\affiliation{\institution{Ant Group}
    \country{}}
\email{colin.zj@antgroup.com}

\author{Yuchang Zhu}
\affiliation{\institution{Sun Yat-sen University}
    \country{}}
\email{zhuych27@mail2.sysu.edu.cn}

\author{Ruofan Wu}
\affiliation{\institution{Ant Group}
    \country{}}
\email{ruofan.wrf@antgroup.com}

\author{Baokun Wang}
\affiliation{\institution{Ant Group}
    \country{}}
\email{yike.wbk@antgroup.com}

\author{Zhen Zhang}
\affiliation{\institution{Ant Group}
    \country{}}
\email{yilue.zz@antgroup.com}

\author{Changhua Meng}
\affiliation{\institution{Ant Group}
    \country{}}
\email{changhua.mch@antgroup.com}

\author{Hong Jin}
\affiliation{\institution{Ant Group}
    \country{}}
\email{jinhong.jh@antgroup.com}

\author{Zibin Zheng}
\affiliation{\institution{Sun Yat-sen University}
    \country{}}
\email{zhzibin@mail.sysu.edu.cn}

\author{Liang Chen}
\authornote{Corresponding author.}
\affiliation{\institution{Sun Yat-sen University}
    \country{}}
\email{chenliang6@mail.sysu.edu.cn}

\renewcommand{\shortauthors}{Li et al.}

\begin{abstract}
    Real-world graphs are typically complex, exhibiting heterogeneity in the global structure, as well as strong heterophily within local neighborhoods. While a growing body of literature has revealed the limitations of common graph neural networks~(GNNs) in handling homogeneous graphs with heterophily, little work has been conducted on investigating the heterophily properties in the context of heterogeneous graphs. To bridge this research gap, we identify the heterophily in heterogeneous graphs using metapaths and propose two practical metrics to quantitatively describe the levels of heterophily. Through in-depth investigations on several real-world heterogeneous graphs exhibiting varying levels of heterophily, we have observed that heterogeneous graph neural networks~(HGNNs), which inherit many mechanisms from GNNs designed for homogeneous graphs, fail to generalize to heterogeneous graphs with heterophily or low level of homophily. To address the challenge, we present \ours, a heterophily-aware HGNN that incorporates both masked metapath prediction and masked label prediction tasks to effectively and flexibly handle both homophilic and heterophilic heterogeneous graphs. We evaluate the performance of \ours on five real-world heterogeneous graph benchmarks with varying levels of heterophily. The results demonstrate that \ours outperforms strong baselines in the semi-supervised node classification task, providing valuable insights into effectively handling more complex heterogeneous graphs.

\end{abstract}




\maketitle

\section{Introduction}
Graphs have become increasingly prevalent in the real world, especially with the rapid development of the World Wide Web (WWW). The interconnected nature of the Web, with webpages linking to each other and users connecting through social networks, naturally lends itself to a graph representation. Graph neural networks (GNNs) have emerged as a powerful framework for analyzing graph-structured data, achieving state-of-the-art performance on various graph learning tasks~\cite{kipf2016semi,velivckovic2017graph,wu2019simplifying}. In recent years, researchers have actively explored the potential of GNNs in handling heterogeneous graphs~\cite{wang2019heterogeneous,hu2020heterogeneous,lv2021we}. Heterogeneous graphs, also known as heterogeneous information networks, are network structures that comprise multiple types of nodes and edges. Each node or edge represents a different type of entity or relationship. The multitude of node and edge types in heterogeneous graphs poses great challenges in exploring rich and diverse semantics on and between nodes.

In response to the challenge of heterogeneity, numerous heterogeneous graph neural networks (HGNNs) have been proposed to address the relevant tasks, including node classification, link prediction, and recommendation. As an extention of GNNs in the context of heterogeneous graphs, HGNNs can learn jointly structural and semantic information in heterogeneous graphs into node representations. One line of research in HGNNs is to define and employ metapaths to model heterogeneous structures~\cite{zhang2019heterogeneous,wang2019heterogeneous,dong2017metapath2vec,fu2020magnn}. Another line of research involves metapath-free paradigms that aggregate messages from a node's local neighborhood like common GNNs but use extra modules to embed semantic information such as node types and edge types into propagated messages~\cite{schlichtkrull2018modeling,zhu2019relation,hu2020heterogeneous,hong2020attention,lv2021we}.

In parallel with the success of GNNs and HGNNs, a growing body of literature~\cite{zhu2020beyond,PeiWCLY20,zhu2021graph,bo2021beyond,zhu2020beyond,yan2022two,LuanHLZZZCP22} has highlighted a significant limitation: common GNNs often exhibit poor performance on heterophilic (particularly homogeneous) graphs, where linked nodes may possess dissimilar labels and attributes (``opposites attract''). While existing HGNNs inherit many mechanisms from GNNs over homogeneous graphs, it remains unclear whether HGNNs can effectively handle heterogeneous graphs with heterophily. Yet, it is non-trivial to properly define the heterophily in heterogeneous graphs due to the heterogeneity.

\begin{figure}[t]
    \centering
    \includegraphics[width=0.6\linewidth]{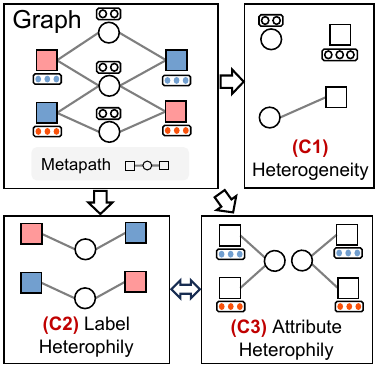}
    \caption{The illustration of three challenges in real-world graphs: C1 - complex structure with heterogeneous types of nodes and edges; C2 \& C3 - heterophilic attributes \& labels that each node type associates.}
    \vspace{-5mm}
    \label{fig:example}
\end{figure}

To bridge this research gap, this paper aims to establish a new foundation by providing a definition of heterophily on heterogeneous graphs using \textit{metapaths}.
As illustrated in Figure~\ref{fig:example}, we identify three major challenges in learning over heterogeneous graphs: heterogeneity, label heterophily, and attribute heterophily.
In this context, heterophily refers to the presence of dissimilarities or differences between two nodes of the same type that are connected by a metapath.
To provide a quantitative description of the levels of heterophily, we introduce two practical metrics: \textit{metapath-based label homophily (MLH)} and \textit{metapath-based Dirichlet energy (MDE)}.
MLH extends the concept of homophily ratio from homogeneous graphs to heterogeneous ones, while MDE is defined based on the Dirichlet energy to assess the smoothness of features in local neighborhoods. Through our empirical study, we observe that HGNNs struggle to generalize to heterogeneous graphs with heterophily (or low/medium level of homophily), as measured by MLH and MDE. In such cases, we also find that even models that ignore the graph structure altogether, such as multilayer perceptrons or MLPs, can outperform HGNNs in many cases.

Motivated by this limitation, we present \ours, a novel heterophily-aware HGNN to effectively and flexibly handle both homophilic and heterophilic heterogeneous graphs. Technically, \ours adopt masked metapath prediction to jointly learn disentangled homophilic and heterophilic representations. Both representations capture different graph signals that facilitate downstream tasks. Additionally, \ours incorporates masked label prediction to enhance message propagation among nodes that exhibit strong label heterophily.

Our contributions are summarized as follows:

\begin{itemize}
    \item \textbf{Heterophily Measures \& Current Limitations:} We introduce two metapath-based metrics, MLH and MDE, which can measure the label and attribute heterophily of heterogeneous graphs, respectively. Our empirical experiments reveal the limitation of HGNNs to learn over heterogeneous graphs with heterophily, which has been largely ignored in the literature due to the lack of well-defined measures.
    \item \textbf{New Model for Heterophily:} We propose \ours, a novel heterophily-aware HGNN in response to the challenge of learning on heterophily yet heterogeneous graphs. By incorporating both disentangled masked graph prediction and masked label prediction tasks, \ours can effectively and flexibly handle both homophilic and heterophilic heterogeneous graphs.
    \item \textbf{Extensive Empirical Evaluation:} We conduct experiments on five real-world heterogeneous graphs, including an industrial-scale graph (13M nodes and 157M edges). \ours demonstrates superior performance compared to strong baselines in the semi-supervised node classification task, shedding light on handling more complex heterogeneous graphs.
\end{itemize}

\section{Related Work}

In this section, we provide a review of relevant literature that is closely related to our work, including HGNNs and graph representation learning over heterophilic graphs.

\subsection{Heterogeneous Graphs Neural Networks}
In the past few years, a broad range of works on HGNNs has emerged to meet the increasing demand for effectively modeling various heterogeneous graphs~\cite{wang2022survey}. According to the way to deal with different semantics, HGNNs can be broadly categorized into metapath-based and metapath-free methods.

Metapath-based HGNNs propagate and aggregate neighbor features using hand-crafted or automatically selected metapaths. For example, Metapath2Vec~\cite{dong2017metapath2vec} adopts metapath-guided random walks to capture the semantic information of heterogeneous nodes.
GTN~\cite{YunJKKK19} assigns learnable weights to different metapaths to automatically learn useful metapaths.
HAN~\cite{wang2019heterogeneous} leverages hierarchical attention to describe node-level and semantic-level structures. As a follow-up work, MAGNN~\cite{fu2020magnn} improves HAN by introducing metapath-based aggregation to learn semantic messages from multiple metapaths.

Metapath-free HGNNs extend message passing and aggregation of GNNs to heterogeneous graphs without manual-designed meaningful metapaths. RGCN~\cite{schlichtkrull2018modeling} and its follow-up work RGAT~\cite{ishiwatari2020relation} propose to capture relation-specific patterns for each edge type with GNNs and then fuse different semantic information together. SHGN~\cite{lv2021we} incorporates a multi-layer graph attention network and extends the edge attention with a learnable edge-type embedding. HGSL~\cite{zhao2021heterogeneous} jointly learns the heterogeneous graph structure and the GNN parameters by exploiting complex interactions.
Inspired by the success of Transformer\cite{VaswaniSPUJGKP17}, HGT~\cite{hu2020heterogeneous} and HINormer~\cite{mao2023hinormer} propose to incorporate the self-attention into the graph-based message passing mechanism for modeling the structural dependencies among heterogeneous nodes.

Despite the progress in HGNNs, few methods explore the heterophily problem in heterogeneous graphs.
We note that a recent work~\cite{guo2023homophily} seeks to address this challenging problem; however, it only scratches the surface by focusing on studying label heterophily.

\subsection{Learning Over Graphs with Heterophily}
Graphs with heterophily, where connected nodes are likely to possess distinct properties or labels, have been receiving growing attention in the research community~\cite{zhu2020beyond,PeiWCLY20,zhu2021graph,bo2021beyond,zhu2020beyond,yan2022two,LuanHLZZZCP22}.
Pei~\textit{et al.}~\cite{PeiWCLY20} first pays attention to this property and provides a metric to measure the homophily level of a graph. Subsequently, Zhu \textit{et al.}~\cite{zhu2020beyond} further investigated the performance degradation of GNNs when learning from graphs with heterophily or non-homophilic characteristics. They proposed $\text{H}_2\text{GCN}$ to enhance learning from the graph structure in the presence of heterophily.
CPGNN~\cite{zhu2021graph} generalizes GNNs to handle graphs with both homophily and heterophily by incorporating a compatibility matrix. FAGCN~\cite{bo2021beyond} introduces an adaptive integration mechanism for low-frequency and high-frequency signals during message passing.
LINKX~\cite{zhu2020beyond} focuses on scaling GNNs to large-scale heterophilic graphs by leveraging MLPs to learn and fuse feature matrix and adjacency matrix information.
Luan~\textit{et al.}~\cite{LuanHLZZZCP22} analyze heterophily from the perspective of post-aggregation node similarity and propose a multi-channel mixing mechanism to extract richer localized information for diverse node heterophily situations. However, the majority of research has been conducted on homogeneous graphs, leaving the heterophily issue largely unexplored in the context of heterogeneous graphs.

\section{Preliminary}
In this section, we introduce the basic concepts and notations of heterogeneous graphs, as well as metrics commonly used for measuring graph homophily/heterophily.

\subsection{Heterogeneous Graphs}

A heterogeneous graph or heterogeneous information network is defined as $\mathcal{G}=\{\mathcal{V}, \mathcal{E}, \mathcal{A}, \mathcal{R}, \phi, \psi\}$, where $\mathcal{V}$ and $\mathcal{E}$ represent the set of nodes and edges, respectively. The graph $\mathcal{G}$ is associated with a node type mapping function $\phi: \mathcal{V} \rightarrow \mathcal{A}$ and correspondingly an edge type mapping function $\psi: \mathcal{E} \rightarrow \mathcal{R}$, where $\mathcal{A}$ denotes the set of possible object types and $\mathcal{R}$ denotes the set of relations (edge types), particularly $|\mathcal{A}| + |\mathcal{R}|>2$.
Note that $\mathcal{G}$ refers to a homogeneous graph when $|\mathcal{A}|=| \mathcal{R}|=1$. In most cases, $\mathcal{G}$ is attributed, meaning that each node $u \in \mathcal{V}$ is associated with an $d$-dimensional attribute vector $x_u \in \mathbb{R}^d$.

\subsection{Metapath in Heterogeneous Graphs}

\nosection{Metapath}
A metapath of length $n$ is denoted as $\mathcal{P} \triangleq A_1 \stackrel{R_1}{\longrightarrow} A_2 \stackrel{R_2}{\longrightarrow} \cdots \stackrel{R_n}{\longrightarrow} A_{n+1}$ (abbreviated as $A_1A_2 \cdots A_{n+1}$), where $A_i \in \mathcal{A}$ and $R_i \in \mathcal{R}$, describing a composite relation $R=R_1R_2\cdots R_n$ between node types $A_1$ and $A_{n+1}$.
For example, a metapath of ``author $\longrightarrow$ paper $\longrightarrow$ author'' in an academic heterogeneous graph indicates the co-author relationship.
Typically, a metapath contains complex semantics and high-order relations, which is considered a vital component for exploring heterogeneous graphs.

\nosection{Relating heterogeneous graphs to homogeneous graphs}
Given a metapath $\mathcal{P}=A_1A_2\cdots A_{n+1}$, we can construct the corresponding metapath induced subgraph $\mathcal{G}_{\mathcal{P}}$, which satisfies that edge $u\in A_1 \rightarrow v \in A_{n+1}$ exists in $\mathcal{G}_{\mathcal{P}}$ if and only if there is at least one length-$n$ path between $u$ and $v$ following the metapath $\mathcal{P}$ in the original graph $\mathcal{G}$. One step further, let $A_1=A_{n+1}$, we have an induced homogeneous graph $\mathcal{G}_{\mathcal{P}}$ built upon a metapath with the end nodes are with the same type.

\subsection{Homophily and Heterophily}
\label{sec:homo_hetero}
\nosection{Homophily metrics}
{
    The homophily metric is a way to measure the extent of correlation between certain types of graph attributes and adjacency structure, with the most widely adopted definition being the edge homophily which is defined as the average agreement over adjacent nodes' label pairs \cite{zhu2020beyond}:
    \begin{align}
        \label{eq:homophily_edge}	\mathcal{H}_\text{edge}\left(\mathcal{G}\right) =\frac{1}{\left|\mathcal{E}\right|} \sum_{\left(u, v\right) \in \mathcal{E}} \mathbbm{1} \left(y_u=y_v\right).
    \end{align}
    The definition Eq.~\eqref{eq:homophily_edge} is defined on the \emph{graph}-level, and might not be ideal for a fine-grained explanation of GNN performance when the underlying task is on the \emph{node}-level, which is also of main interest in this paper. Therefore we additionally adopt another useful definition that is essentially a reweighted version of Eq.~\eqref{eq:homophily_edge} via considering the node-wise average of averaged neighborhood agreement, which is also referred to as node homophily \cite{luan2022revisiting, luan2023graph}:
    \begin{align}
        \label{eq:homophily_node}
        \mathcal{H}_\text{node}\left(\mathcal{G}\right) = \frac{1}{|\mathcal{V}|} \sum_{u \in \mathcal{V}}\mathcal{H}(u) \triangleq \frac{1}{|\mathcal{V}|} \sum_{u \in \mathcal{V}} \frac{\sum_{v \in \mathcal{N}_u}\mathbbm{1} \left(y_u=y_v\right) }{|\mathcal{N}_u|}.
    \end{align}
    An advantage of the definition Eq.~\eqref{eq:homophily_node} is that it enables assessing the correlation of node-wise tasks with local homophily metrics $\mathcal{H}(u)$. \par
    Despite their popularity, a notable caveat of label-related homophily definitions is that they may be ill-defined when the underlying graph is label-scarce, which is often encountered in practice. There have been recent proposals that generalize the definition of homophily beyond label information \cite{jin2022raw, luan2022revisiting}, which treat homophily as a measure of attribute smoothness over the underlying graph. Motivated by such generalizations, we propose to use \emph{Dirichlet energy} regarding node features as another measure of homophily. Specifically, we begin with the definition in \cite{maskey2023fractional}:
    \begin{equation}
        \label{eq:energy}
        \mathscr{E}_\text{edge}(\mathcal{G})=\frac{1}{4} \sum_{(u, v) \in \mathcal{E}} \left\|\frac{\mathrm{x}_u}{\sqrt{|\mathcal{N}_u|}}-\frac{\mathrm{x}_v}{\sqrt{|\mathcal{N}_v|}}\right\|_2^2.
    \end{equation}
    Analogously, we may generalize definition Eq.~\eqref{eq:energy} to its node-level counterpart $\mathscr{E}_\text{node}$ by computing the node-wise average of local energies $\mathscr{E}(u)$, following similar philosophy as described in Eq.~\eqref{eq:homophily_node}. We omit it in this context for the sake of simplicity. A smaller Dirichlet energy is regarded as indicating stronger evidence of attribute-level homophily.
    \footnote{By saying \emph{small}, we are implicitly assuming that the scale of $\mathscr{E}_\text{edge}(\mathcal{G})$ or $\mathscr{E}_\text{node}(\mathcal{G})$ is controlled within a proper range, which might not be the case for energy-based definitions unless the feature matrix is appropriately normalized \cite{maskey2023fractional}. In the subsequent explorations, we will empirically verify that the actual scale of these metrics do not affect our findings. Therefore we postpone a more detailed discussion regarding the scale of Dirichlet energies to appendix \ref{sec:discussion-energy}. }
    The introduction of two notions of homophily is somewhat necessary, since the performance of GNN is affected by the distributions of both attributes and labels, which are separately captured by the two metrics introduced.
}

\nosection{Heterophily $\neq$ heterogeneity}
In line with the focus of this work, we remark that the terms heterophily and heterogeneity are distinct concepts.
Heterophily, formally defined, refers to the tendency of nodes with different attributes or labels to be connected in a graph. It emphasizes the existence of connections between nodes with contrasting characteristics. On the other hand, heterogeneity refers to the diversity or variability of nodes and edges within the graph. It is worth noting that a graph can be both heterophilic and heterophilic, regardless of its heterogeneity.

\section{Empirical Investigation on Heterophily}

In this section, we perform empirical investigations on how representative HGNNs behave on heterogeneous graphs with heterophilic labels and attributes to support our claims.

\subsection{Metapath-based Homophily Metrics}
Recall that in Section~\ref{sec:homo_hetero}, we introduced two metrics to measure the label and attribute homophily of a graph, respectively. However, these metrics were formally defined for homogeneous graphs and cannot be directly applied to heterogeneous graphs. This limitation has motivated us to introduce two metapath-based homophily metrics that extend the definition of homophily ratio and Dirichlet energy to heterogeneous graphs.
Formally, for a given node type $A\in \mathcal{A}$, we define the \textit{metapath-based label homophily} (MLH) and \textit{metapath-based Dirichlet energy} (MDE) as follows:
\begin{equation}
    \label{eq:metrics}
    \begin{aligned}
        \text{MLH}_A\left(\mathcal{G}\right) & =\frac{1}{|\mathscr{P}|}\sum_{\mathcal{P} \in \mathscr{P}}\mathcal{H}(\mathcal{G}_\mathcal{P}), \\
        \text{MDE}_A\left(\mathcal{G}\right) & =\frac{1}{|\mathscr{P}|}\sum_{\mathcal{P} \in \mathscr{P}}\mathscr{E}(\mathcal{G}_\mathcal{P}),
    \end{aligned}
\end{equation}
where $\mathscr{P}=\{A \cdots A_i\cdots A|A_i \in \mathcal{A}\}$ represents the set of all possible metapaths starting from node type $A$ and ending at node type $A$. Both $\mathcal{H}$ and $\mathscr{E}$ can be either edge-level or node-level metrics.
Since the set of all possible metapaths $\mathscr{P}$ can potentially be infinitely large, we introduce an additional constraint where only \textit{length-2 metapaths} are considered.
Typically, a larger MLH or a smaller MDE suggests a stronger homophily of a heterogeneous graph.

\begin{figure}[t]
    \centering
    \subfigure[MLH]{
        \includegraphics[width=0.45\linewidth]{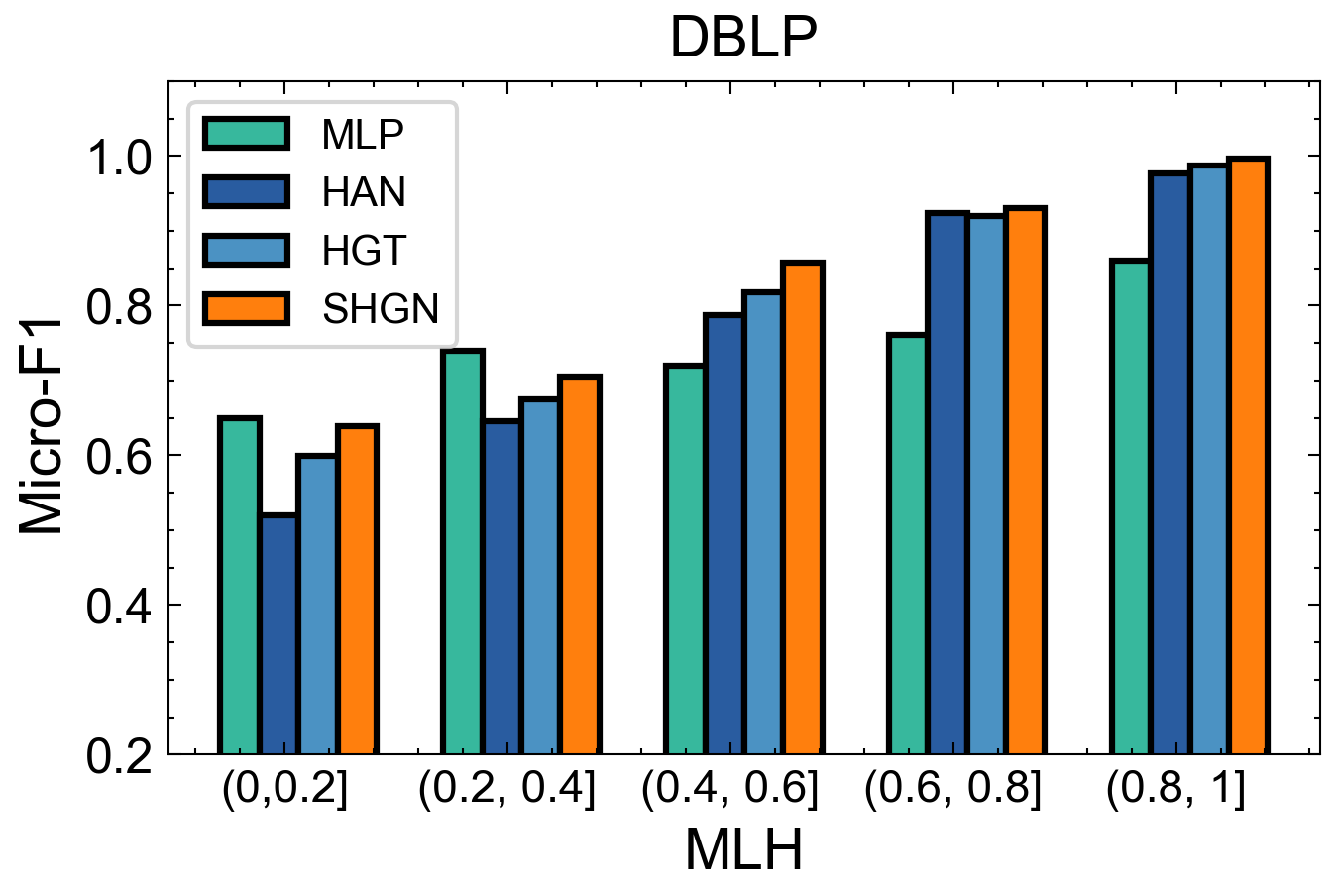}
        \includegraphics[width=0.45\linewidth]{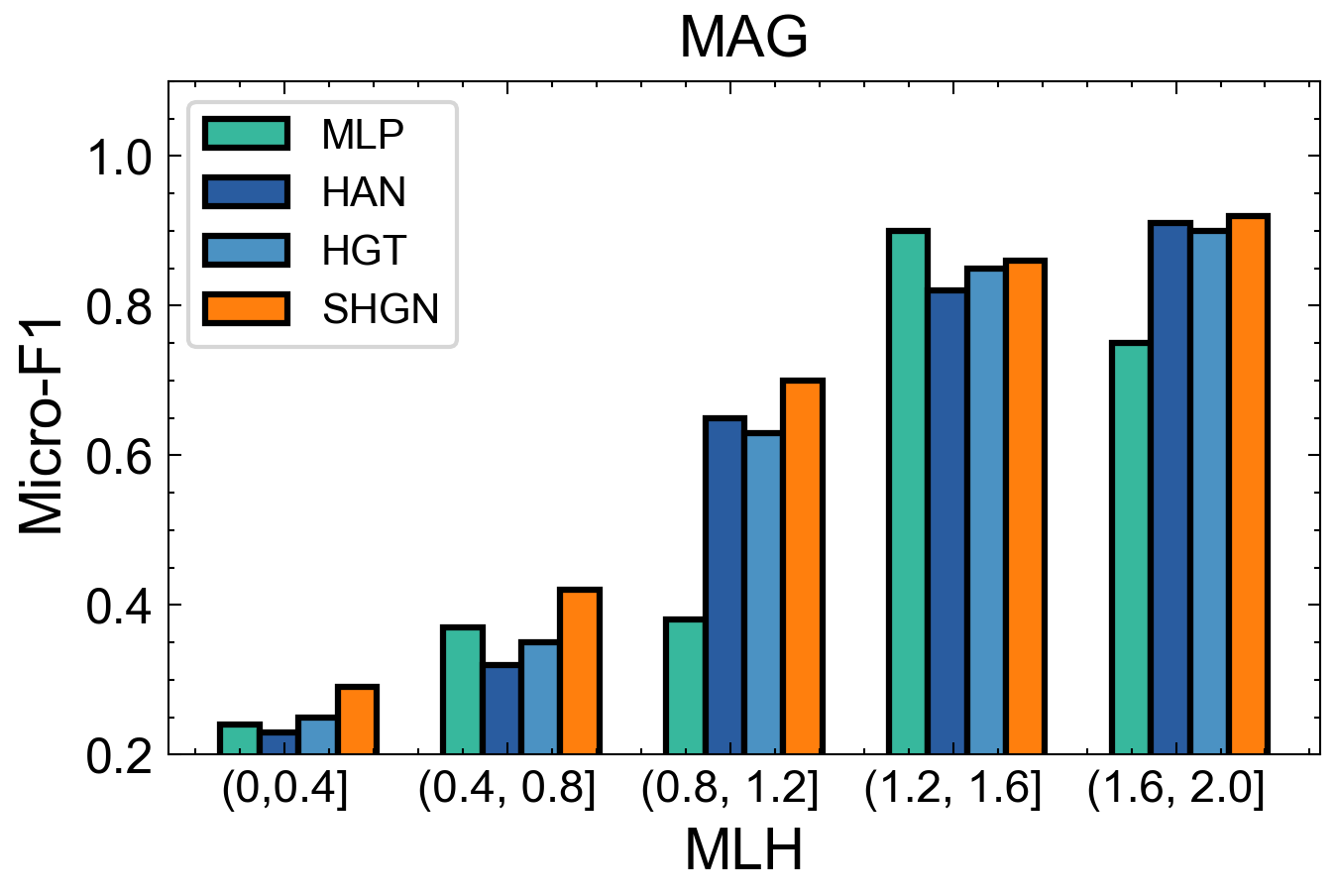}}
    \subfigure[MDE]{
        \includegraphics[width=0.45\linewidth]{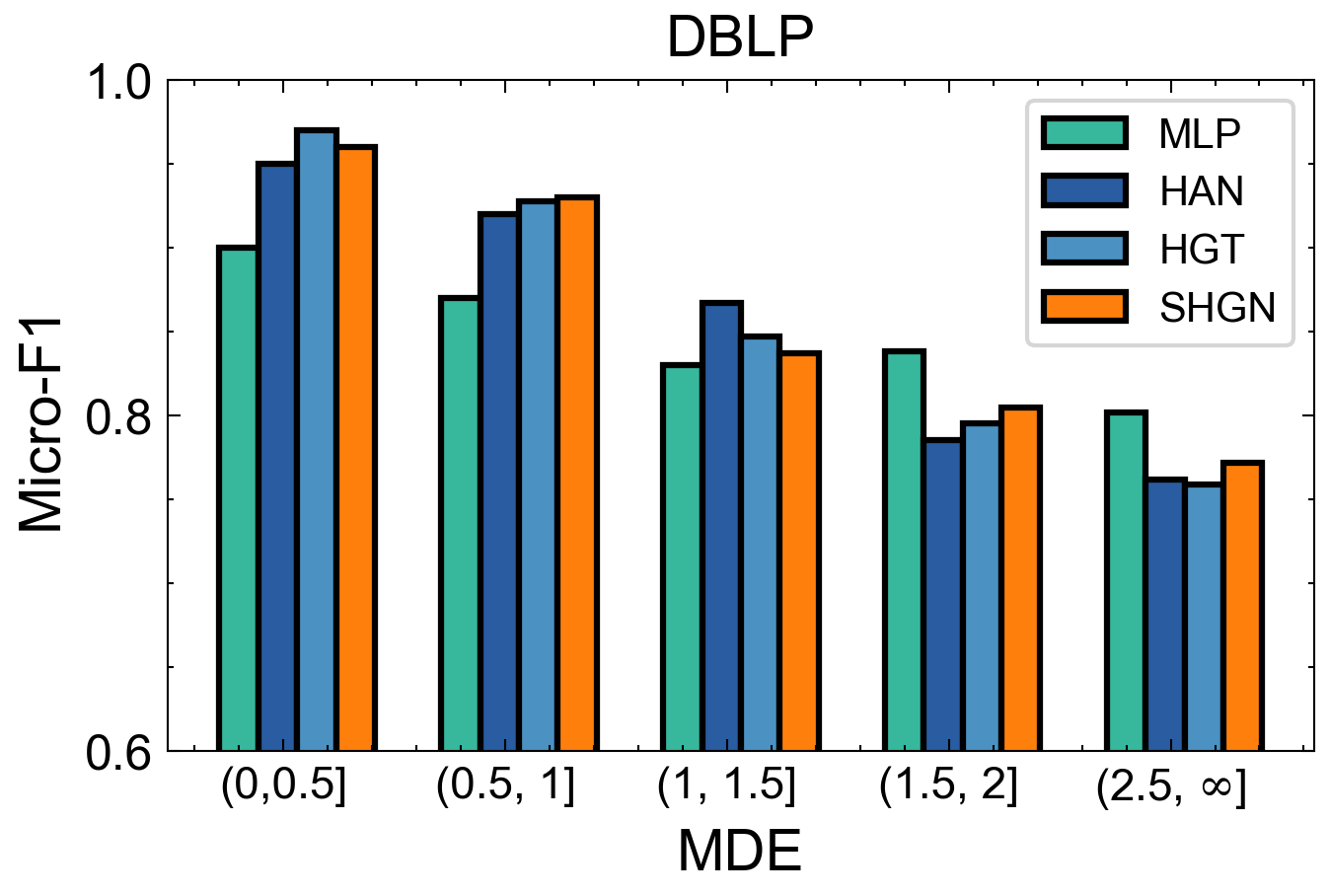}
        \includegraphics[width=0.45\linewidth]{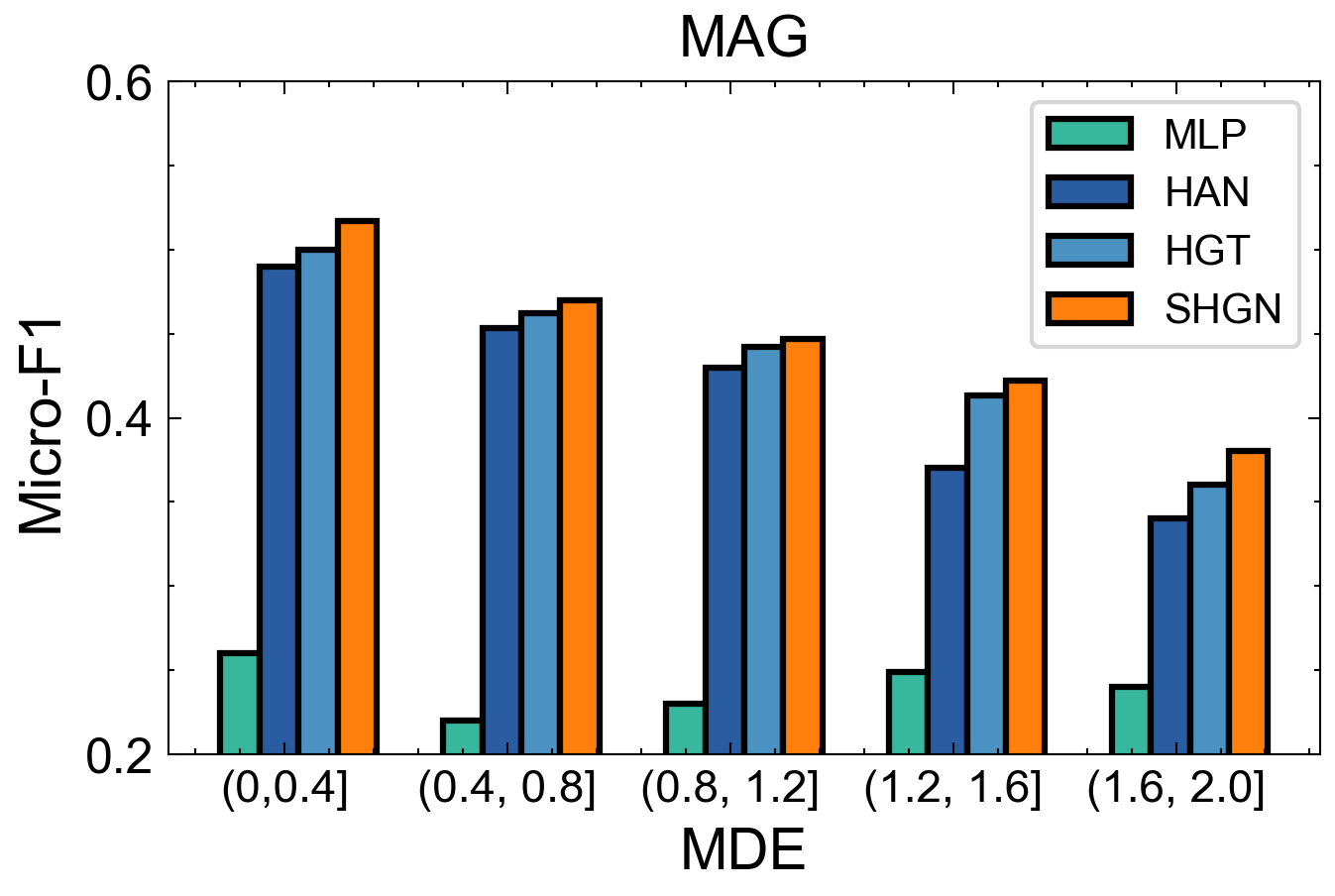}
    }
    \vspace{-4mm}
    \caption{Performance of HGNNs in terms of nodes with different MLH and MDE, respectively.}
    \label{fig:homophily}
\end{figure}

\begin{figure*}[t]
    \centering
    \includegraphics[width=.85\linewidth]{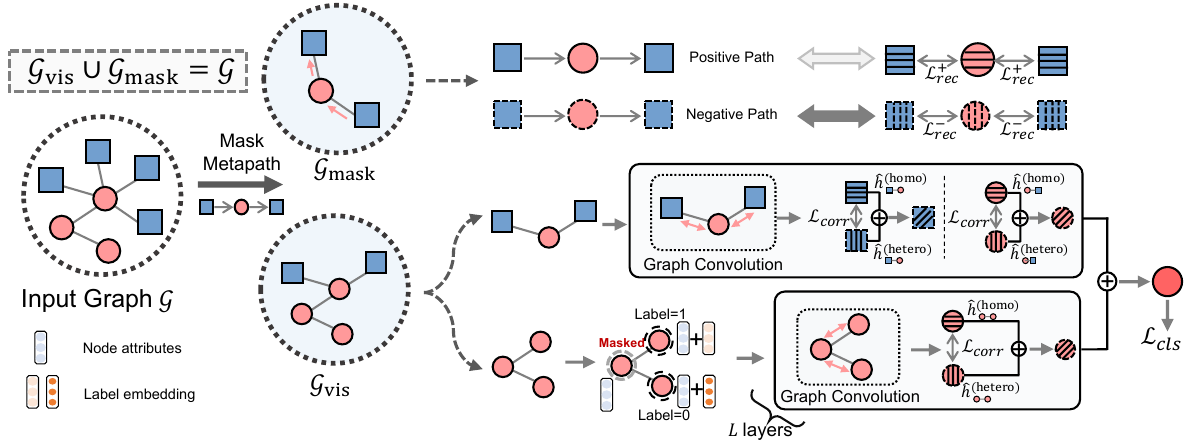}
    \caption{The overall framework of \ours. \ours is a semi-supervised learning framework with two auxiliary tasks: (i) masked metapath prediction and (ii) masked label prediction. In particular, the goal of \ours is to
        learn disentangled homophilic and heterophilic representations from each individual view of the heterogeneous graph by predicting the masked metapaths and the masked node labels.}
    \label{fig:framework}
\end{figure*}

\subsection{Empirical study}
\label{sec:empiricals}
\nosection{Experimental setup}
To investigate how HGNNs behave on heterophilic graphs, we design semi-supervised node classification experiments on two heterogeneous graph datasets, DBLP and MAG, which have different scales and heterophily. Dataset statistics are listed in Table~\ref{tab:dataset}. The experiments are performed on three representative HGNNs covering metapath-based and metapath-free architectures, including HAN~\cite{wang2019heterogeneous}, HGT~\cite{hu2020heterogeneous}, and SHGN~\cite{lv2021we}. All HGNNs follow the hyperparameter settings in their original papers.
In addition, we include a multilayer perceptron (MLP) as a baseline for comparison, which does not utilize any structural information from the graph.
    {We calculate the MLH and MDE over disjoint subgraphs that are partitioned by local node-level homophily metrics, and examine how these models perform on nodes with different levels of heterophily}
We report the average performance of Micro-F1 scores over 5 runs. The Micro-F1 scores on nodes with different MLH and MDE on DBLP and MAG are illustrated in Figure~\ref{fig:homophily}.

\nosection{Observations}
From Figure~\ref{fig:homophily}, we can make the following observations: (i) Both MLH and MDE metrics demonstrate strong correlations with the performance of HGNNs, revealing a large discrepancy between nodes with varying MLH and MDE values. Specifically, HGNNs tend to achieve better performance on nodes that exhibit strong homophily in their local structure, as indicated by larger values of MLH or smaller values of MDE. (ii) HGNNs fail to generalize effectively to nodes with strong heterophily. In particular, a simple MLP can even outperform HGNNs on such nodes in most cases.
(iii) In the DBLP dataset, which exhibits high label homophily, the performance gaps between nodes with different MLH values are relatively smaller. Conversely, in the MAG dataset, the performance gaps between nodes with different MLH values are larger. We can observe a similar trend revealed by the MDE metric as well.

In conclusion, both MLH and MDE provide valuable insights for investigating the heterophilic property in heterogeneous graphs. The heterophily issue poses a significant challenge for HGNNs, yet it has been largely overlooked in current literature. This highlights the need for further research efforts to effectively address the heterophily issue.

\section{Methodology}
In this section, we present \ours as a heterophily-aware representation learning framework for heterogeneous graphs.
As shown in Figure~\ref{fig:framework}, \ours first extends the message passing scheme from homogeneous graphs to heterogeneous ones by incorporating a multi-view fusion perspective (Section~\ref{sec:multi_view}). Then, \ours incorporates two key designs to address the challenges posed by label and attribute heterophily, including (i) disentangled masked metapath prediction (Section~\ref{sec:dis_mmp}) and (ii) masked label prediction (Section~\ref{sec:mask_lp}).
In what follows, we will give the details of the proposed \ours framework.

\subsection{Multi-view Graph Fusion}
\label{sec:multi_view}

Representation learning on heterogeneous graphs can be approached from a multi-view graph fusion perspective, which involves performing message passing within individual graph views and simultaneously conducting graph fusion across multiple views.
Specifically, a heterogeneous graph composed of $|\mathcal{R}|$ types of relations can be regarded as a multiplex graph, consisting of $|\mathcal{R}|$ individual views.
Each view represents a bipartite graph that focuses on a specific relationship type between nodes.
For each graph view formulated by a relation $\psi(u,v)$ with $u$ as the source node and $v$ as the target node, we perform the well-established \textit{message passing} scheme in which each representation of node $u$ is computed recursively by aggregating representations from its immediate neighbors that are of a specific type $\phi(v)$.
Formally, given $L$ graph convolution layers, the updating process of the $l$-th layer in each graph view could be formulated as:
\begin{equation}
    \label{eq:agg}
    \begin{aligned}
        h_u^{(l)}(\psi(u,v)) & =\textsc{Com}\left(\left\{\textsc{Aggr}\left(\left\{h_{v}^{(l-1)}\left(\phi(v)\right): v \in \mathcal{N}_u(\phi(v))\right\}\right)\right\}\right), \\
    \end{aligned}
\end{equation}
where $\textsc{Aggr}(\cdot)$ is the \textit{aggregation} function that aggregates representations from the neighbors (in some way—e.g., average), and $\textsc{Com}(\cdot)$ denotes the \textit{combination} of aggregated features from a central node and its neighbors. Both $\textsc{Aggr}$ and $\textsc{Com}$ may be followed by a non-linear transformation with learnable parameters. $\mathcal{N}_u(\phi(v))$ is the set of $u$'s neighbors that is of type $\phi(v)$. $h_u^{(l)}(\cdot)$ is the (aggregated) representation of node $u$ at the $l$-th layer and initially $h_u^{(0)}=x_u$.

Then, the representations from different views that correspond to the same node type are fused or merged together to obtain the node representation:
\begin{equation}
    \begin{aligned}
        h_u^{(l)}(\phi(u)) & = \textsc{Fuse}\left(\left\{\hat{h}_u^{(l)}(\psi(u,v))| \psi(u, v)\in \mathcal{R}\right\} \right),
    \end{aligned}
    \label{eq:fuse}
\end{equation}
where $\textsc{Fuse}(\cdot)$ denotes the \textit{fusion} on representations of the same semantic learned from different views. Typically, $\textsc{Fuse}(\cdot)$ is implemented as one of the following functions: $\textsc{Mean}$, $\textsc{Sum}$, or $\textsc{Concat}$. In particular, RGCN~\cite{schlichtkrull2018modeling} is the special case with $\textsc{Sum}$ the grouping function.
Note that each graph view in the heterogeneous graph has its own set of graph convolution layers with learnable parameters. This means that the overall model, with a depth of $L$, consists of $L \times |\mathcal{R}|$ layers.

\subsection{Disentangled Masked Metapath Prediction}
\label{sec:dis_mmp}
Early work on learning with heterophilic graphs has demonstrated that low-frequency information, specifically homophilic signals, plays a crucial role in the success of Graph Neural Networks (GNNs)~\cite{bo2021beyond, wu2019simplifying}. Moreover, heterophilic signals, which capture the difference between nodes, are also vital for effectively handling heterophilic graphs and mitigating the oversmoothing problem. Although both homophilic and heterophilic signals are important, learning and determining the appropriate signals to use from a graph can be challenging due to the complex correlation between tasks and different types of information.

In this work, we resort to disentangled representation learning, which has been widely used to learn underlying representations that disentangle hidden factors in the graph structure.
Our goal is to decouple the \textit{homophilic} and \textit{heterophilic} representations from each individual graph view and explicitly represent the graph with the two disentangled representations:
\begin{equation}
    \hat{h}_u(\psi(u,v)) \triangleq \hat{h}_u^\text{(homo)} + \hat{h}_u^\text{(hetero)},
\end{equation}
where $\hat{h}_u^\text{(homo)}$ and $\hat{h}_u^\text{(hetero)}$ are homophilic and heterophilic representations of node $u$, respectively. Here $\hat{h}_u^\text{(homo)}$ and $\hat{h}_u^\text{(hetero)}$ are learned from two different channels of the same graph convolution layer, as stated in Eq.~\eqref{eq:agg}.
For the sake of notation simplicity, we omit the relation $\psi(u,v)$ from $\hat{h}_u^{(\cdot)}(\psi(u,v))$ hereafter. Particularly, $\hat{h}^\text{(homo)}$ and $\hat{h}^\text{(hetero)}$ should exhibit two \textbf{properties}:
\begin{itemize}
    \item \textbf{(P1)}: $\hat{h}^\text{(homo)}$ and $\hat{h}^\text{(hetero)}$ are disentangled from each other. This means that they are explicitly separated and represent distinct aspects of the graph's information.
    \item \textbf{(P2)}: $\hat{h}^\text{(homo)}$ captures homophilic signals in the node's local neighborhood, while $\hat{h}^\text{(hetero)}$ captures heterophilic signals. In other words, $\hat{h}^\text{(homo)}$ focuses on encoding the similarities and community structure within a node's vicinity, while $\hat{h}^\text{(hetero)}$ emphasizes the differences and dissimilarities between nodes.
\end{itemize}

To encourage node disentanglement \textbf{(P1)} between latent representations $\hat{h}^\text{(homo)}$ and $\hat{h}^\text{(hetero)}$, we have to enforce the statistical independence between them. Obviously, the minimization of the correlation between $\hat{h}^\text{(homo)}$ and $\hat{h}^\text{(hetero)}$ could be used to achieve the disentanglement between homophilic and heterophilic representations. In this work, we adopt a Pearson's correlation coefficient as the correlation loss between $\hat{h}^\text{(homo)}$ and $\hat{h}^\text{(hetero)}$, i.e.,
\begin{equation}
    \label{eq:corr}
    \mathcal{L}_\text{corr}=\sum_{u \in \mathcal{V}}^{} \frac{\left|\textsc{Cov}\left(\hat{h}_u^\text{(homo)}, \hat{h}_u^\text{(hetero)}\right)\right|}{\sqrt{\textsc{Var}\left(\hat{h}_u^\text{(homo)}\right)} \sqrt{\textsc{Var}\left(\hat{h}_u^\text{(hetero)}\right)}},
\end{equation}
where $\textsc{Cov}(\cdot, \cdot)$ and $\textsc{Var}(\cdot)$ indicate covariance and variance operations, respectively.

Although the goal of disentangling $\hat{h}^\text{(homo)}$ and $\hat{h}^\text{(hetero)}$ is achieved by minimizing the correlation loss $\mathcal{L}_\text{corr}$, it is necessary to introduce an additional task to assign semantic meaning to these representations \textbf{(P2)}. In other words, we aim to associate homophily and heterophily with $\hat{h}^\text{(homo)}$ and $\hat{h}^\text{(hetero)}$, respectively. In this work, we extend the homophily assumption from homogeneous graphs\footnote{In literature, it is fundamentally assumed that nodes with similar attributes, characteristics, or properties are more likely to be connected in the graph.} to heterogeneous graphs by leveraging metapaths, i.e.,
\begin{center}
    \textit{Nodes connected by a metapath are more likely to share homophilic properties, such as classes or attributes.}
\end{center}

Based on the above assumption, it is natural to learn $\hat{h}^\text{(homo)}$ by enforcing the proximity between nodes and their local neighborhoods. There are several alternative tasks available in the literature, but in this work, our focus is on masked graph autoencoding. Masked graph autoencoding has emerged as a promising self-supervised task and has demonstrated remarkable performance in various graph learning tasks~\cite{maskgae,hgmae}.

Inspired by the recent work MaskGAE\cite{maskgae}, which focuses on masked edge/path reconstruction in homogeneous graphs, our work shifts its focus to masked graph modeling in heterogeneous graphs. In this subsection, we introduce a principled task that utilizes masked metapath prediction to learn semantic homophilic representations.
Given a set of pre-defined metapaths $\mathcal{P}^*=\{\mathcal{P}_1,\mathcal{P}_2,\ldots\}$, the input heterogeneous graph is masked by a set of adjacent paths
$\rho^+=\{u_1u_2\cdots u_\text{n}|\phi(u_1)\phi(u_2)\cdots \phi(u_\text{n})\in \mathcal{P}^* \wedge (u_i,u_j) \in \mathcal{E}\}$.
Let $\mathcal{G}_\text{mask}$ denote the graph masked from $\mathcal{G}$ (with $\rho^+$ masked) and $\mathcal{G}_\text{vis}$ the remaining visible graph. our goal is to learn $h^\text{(homo)}$ by reconstructing $\mathcal{G}_\text{mask}$ using the remaining visible structure $\mathcal{G}_\text{vis}$.
Given that two nodes disconnected would potentially share heterophilic information, we can also learn $h^\text{(hetero)}$ by reconstructing the negative metapaths $\rho^-=\{u_1u_2\cdots u_\text{n}|\phi(u_1)\phi(u_2)\cdots \phi(u_\text{n})\in \mathcal{P}^* \wedge (u_i,u_j) \notin \mathcal{E}\}$.
Overall, we optimize the following binary cross-entropy loss:
\begin{equation}
    \label{eq:rec}
    \begin{aligned}
         & \mathcal{L}_\text{rec}^+ = \frac{1}{|\rho^+|}\sum_{(u, v)\in \rho^+}\log f_\theta(h_u^\text{(homo)}, h_v^\text{(homo)}),     \\
         & \mathcal{L}_\text{rec}^- = \frac{1}{|\rho^-|}\sum_{(u, v)\in \rho^-}\log f_\theta(h_u^\text{(hetero)}, h_v^\text{(hetero)}), \\
         & \mathcal{L}_\text{rec} = - \left(\mathcal{L}_\text{rec}^++\mathcal{L}_\text{rec}^-\right),
    \end{aligned}
\end{equation}
where $f_\theta$ is a standard 2-layer MLP parameterized by $\theta$, which is defined as:
\begin{equation}
    f_\theta(h_u, h_v) = \mathbf{W}_1 \cdot  \textsc{ReLU}(\mathbf{W}_0 \cdot (h_u \circ h_v)), \quad \theta=\{\mathbf{W}_0, \mathbf{W}_1\},
\end{equation}
where $\circ$ denotes the element-wise product, and $\mathbf{W}_0$ and $\mathbf{W}_1$ represent learnable matrices of MLP. We omit the bias terms in the MLP for simplicity, although they can be easily incorporated if desired.
By predicting the missing metapaths within the graph, our model is encouraged to capture and understand the underlying relationships and patterns between nodes of different types.

\nosection{Comparison with prior work}
Previous disentangled techniques typically adopt a multichannel graph convolutional layer to promote node disentanglement among diverse latent representations. These methods intend to enforce independent capture of isolated hidden factors by each dimension of the representation in each channel. However, channel-wise or dimension-wise disentanglements might not be efficient in training, especially as the number of channels and dimensions grows. Consequently, disentangled representation learning struggles to scale effectively to large and heterogeneous graphs. In contrast, we present a simple yet effective disentangled way utilizing only two independent channels, i.e., homophilic and heterophilic channels.
One step further, we avoid the need for dimension-wise disentanglement through decorrelation and the masked metapath prediction task. This enables \ours to overcome the scalability limitation often encountered when dealing with large graphs.

\subsection{Masked Label Prediction}
\label{sec:mask_lp}

While disentangled representation learning with masked metapath prediction is effective in handling heterogeneous graphs with heterophily, particularly in node attributes, it does not directly leverage label information during message propagation. This limitation may potentially lead to unsatisfied performance, especially in graphs with strong label heterophily.

Inspired by \cite{ShiHFZWS21}, we resort to label propagation algorithm~\cite{ZhouBLWS03} to solve the label heterophily issue.
Label propagation (LP) is one of the classic message passing algorithms in semi-supervised node classification, which predicts a class for each node by propagating the known labels
in the graph.
The classic LP typically assumes that two connected nodes are more likely to have the same class, and thus iteratively propagate the class labels from labeled nodes to their unlabeled neighbors.
To adapt LP to graphs with heterophily directly, several methods have been explored in the literature, such as employing pseudo labels~\cite{HanLMTSLT23}, compatibility matrix~\cite{ZhongIP22}, or an additional learnable weight matrix~\cite{WangJWHH22}. Although LP can be effective in leveraging the graph structure to infer labels for unlabeled nodes, it is typically used as a standalone component rather than being explicitly integrated into the message passing design.

In this work, we propose to incorporate the partially observed labels into message passing by propagating both ground truth labels and node features through the given graph.
Given partially observed class labels, denoted as $\mathcal{Y}_\text{train}=\{y_0, y_1,\ldots\}$, where each $y_i\in \mathbb{R}^{C\times 1}$ represents a one-hot vector encoding the class label of a labeled node.
We first map the one-hot label vectors $\mathcal{Y}_\text{train}$ into a continuous space using a linear projection. Next, we add the mapped labels into the node features as follows:
\begin{equation}
    \hat{x}_u = x_u + \mathbf{W}_y\cdot y_u,
\end{equation}
where $\mathbf{W}_y$ is a learnable weight matrix that embeds $\mathcal{Y}_\text{train}$ into the same space as node features. By combining the label information with the existing features, we create augmented node representations that capture both the original features and the label information.

To prevent the leakage of label information during training, we adopt a masked label prediction strategy inspired by masked graph autoencoding~\cite{maskgae}. During each training round, we randomly mask a portion of node labels following a specific distribution, e.g., Bernoulli distribution:
\begin{equation}
    \label{eq:lp}
    \begin{aligned}
        m         & \sim \text{Bernoulli}(p),                \\
        \hat{x}_u & = x_u + \mathbf{W}_y\cdot (y_u \circ m),
    \end{aligned}
\end{equation}

where $0<p\leq 1$ is the masking ratio for the node labels.
In this way, the augmented node representations are propagated through the graph, allowing the label information to be shared and refined across neighboring nodes without leakage.
During the inference stage, we set the masking ratio $p$ to 0, which ensures that the model can leverage the complete label information $\mathcal{Y}_\text{train}$ to guide its predictions.

\subsection{Overall Objection Function}
As a supervised learning framework, \ours follows the standard semi-supervised node classification pipeline and employs cross-entropy loss for optimization:
\begin{equation}
    \label{eq:cls}
    \mathcal{L}_\text{cls} = \sum_{v\in \mathcal{V}_\text{tr}}\textsc{CrossEnt}(\tilde{y}_u, y_u),
\end{equation}
where $\textsc{CrossEnt}$ is the cross-entropy loss and $\mathcal{V}_\text{tr}$ is the set of labeled training nodes. Note that, for multi-label node classification, we can employ binary cross-entropy as the optimization loss.

Integrating the correlation loss in Eq.\eqref{eq:corr}, the reconstruction loss in Eq.\eqref{eq:rec}, with the semi-supervised node classification loss Eq.\eqref{eq:cls}, the overall objective function of \ours is formulated as:
\begin{equation}
    \label{eq:obj}
    \mathcal{J} = \mathcal{L}_\text{cls} + \alpha\mathcal{L}_\text{corr} + \beta\mathcal{L}_\text{rec},
\end{equation}
where $\alpha\geq0$ and $\beta\geq0$ are non-negative hyperparameters trading off three terms. To help better understand the proposed framework, we provide the detailed algorithm for training \ours in Appendix~\ref{sec:appendix-algo}.

\begin{table}[t]
    \caption{Dataset Statistics. MLH and MDE are calculated over the target node type and darker \textcolor{blue}{color} indicates stronger heterophily in terms of labels or attributes.}
    \label{tab:dataset}
    \begin{threeparttable}
        \resizebox{\linewidth}{!}
        {\begin{tabular}{l|ccccc}
                \toprule
                \textbf{}                       & \textbf{DBLP}       & \textbf{ACM}        & \textbf{IMDB}       & \textbf{MAG}        & \textbf{RCDD}       \\
                \midrule
                \textbf{\#Nodes}                & 26,128              & 10,942              & 21,420              & 1,939,743           & 13,806,619          \\
                \textbf{\#Edges}                & 239,566             & 547,872             & 86,642              & 21,111,007          & 157,814,864         \\
                \textbf{\#Node Types}           & 4                   & 4                   & 4                   & 4                   & 7                   \\
                \textbf{\#Edge Types}           & 6                   & 8                   & 6                   & 4                   & 7                   \\
                \textbf{\#Classes}              & 4                   & 3                   & 5                   & 349                 & 2                   \\
                \textbf{Tr. nodes}              & 30\%                & 30\%                & 28\%                & 94\%                & 0.7\%               \\
                \textbf{Target}                 & author              & paper               & movie               & paper               & item                \\
                \midrule
                \textbf{MLH$^\dagger$ (Label)}  & \colorbox{c1}{0.87} & \colorbox{c2}{0.60} & \colorbox{c4}{0.24} & \colorbox{c5}{0.21} & \colorbox{c3}{0.45} \\
                \textbf{MDE$^\dagger$ (Attri.)} & \colorbox{c5}{0.80} & \colorbox{c4}{0.74} & \colorbox{c3}{0.59} & \colorbox{c1}{0.04} & \colorbox{c2}{0.16} \\
                \bottomrule
            \end{tabular}
        }
        \begin{tablenotes}\footnotesize
            \item $^\dagger$Both MLH and MDE here are calculated on the target node type.
        \end{tablenotes}
    \end{threeparttable}

\end{table}

\begin{table*}[t]
    \caption{Quantitative results (\%) on the semi-supervised node classification task. The results are averaged over five runs.}
    \begin{threeparttable}
        \resizebox{\linewidth}{!}
        {\begin{tabular}{lcccccccc}
                \toprule
                               & \multicolumn{2}{c}{ DBLP }   & \multicolumn{2}{c}{ IMDB }   & \multicolumn{2}{c}{ACM}      & \multicolumn{2}{c}{ MAG }                                                                                                                                 \\
                \cmidrule{2-9} & Macro-F1                     & Micro-F1                     & Macro-F1                     & Micro-F1                     & Macro-F1                      & Micro-F1                     & Macro-F1                     & Micro-F1                     \\
                \midrule
                GCN            & 90.84 $\pm$ 0.32             & 91.47 $\pm$ 0.34             & 57.88 $\pm$ 1.18             & 64.82 $\pm$ 0.64             & 92.17 $\pm$ 0.24              & 92.12 $\pm$ 0.23             & 25.14 $\pm$ 0.33             & 47.26 $\pm$ 0.36             \\
                GAT            & 91.05 $\pm$ 0.76             & 91.73 $\pm$ 0.50             & 58.94 $\pm$ 1.35             & 64.86 $\pm$ 0.43             & 92.26 $\pm$ 0.94              & 92.19 $\pm$ 0.93             & 22.94 $\pm$ 0.49             & 43.79 $\pm$ 0.24             \\
                LINKX          & 75.05 $\pm$ 1.45             & 77.78 $\pm$ 1.59             & 58.98 $\pm$ 0.45             & 62.03 $\pm$ 0.41             & 89.91 $\pm$ 1.14              & 89.69 $\pm$ 1.09             & 14.63 $\pm$ 0.56             & 33.43 $\pm$ 0.22             \\
                FAGCN          & 82.40 $\pm$ 0.28             & 83.08 $\pm$ 0.28             & 63.68 $\pm$ 0.56             & 67.49 $\pm$ 0.25             & 89.27 $\pm$ 0.62              & 89.34 $\pm$ 0.78             & 16.87 $\pm$ 0.41             & 37.99 $\pm$ 0.95             \\
                \midrule
                RGCN           & 91.52 $\pm$ 0.50             & 92.07 $\pm$ 0.50             & 61.26 $\pm$ 0.33             & 65.21 $\pm$ 0.73             & 91.95 $\pm$ 0.44              & 91.75 $\pm$ 0.35             & 27.01 $\pm$ 0.21             & 48.80 $\pm$ 0.24             \\
                HAN            & 91.67 $\pm$ 0.49             & 92.05 $\pm$ 0.62             & 57.74 $\pm$ 0.96             & 64.63 $\pm$ 0.58             & 90.89 $\pm$ 0.43              & 90.79 $\pm$ 0.43             & 8.94 $ \pm 0$.16             & 26.76 $\pm$ 0.36             \\
                RGAT           & 92.61 $\pm$ 0.48             & 93.15 $\pm$ 0.49             & 57.85 $\pm$ 0.58             & 62.79 $\pm$ 0.70             & 90.03 $\pm$ 0.56              & 90.40 $\pm$ 0.54             & 24.76 $\pm$ 0.47             & 45.29 $\pm$ 0.40             \\
                HGT            & 93.01 $\pm$ 0.23             & 93.49 $\pm$ 0.25             & 63.00 $\pm$ 1.19             & 67.20 $\pm$ 0.57             & 91.12 $\pm$ 0.76              & 91.15 $\pm$ 0.71             & \underline{27.87 $\pm$ 0.30} & \underline{49.19 $\pm$ 0.63} \\
                SHGN           & \underline{94.01 $\pm$ 0.24} & 94.20 $\pm$ 0.31             & 63.53 $\pm$ 1.26             & 67.36 $\pm$ 0.57             & \underline{ 93.42 $\pm$ 0.44} & \underline{93.35 $\pm$ 0.45} & 22.61 $\pm$ 0.40             & 43.68 $\pm$ 0.71             \\
                HINormer       & 93.90 $\pm$ 0.17             & \underline{94.32 $\pm$ 0.15} & \underline{64.30 $\pm$ 0.92} & \underline{67.62 $\pm$ 0.52} & 92.23 $\pm$ 0.27              & 92.15 $\pm$ 0.28             & 25.80 $\pm$ 0.41             & 47.57 $\pm$ 0.61             \\
                \midrule
                \ours          & \sota{94.03 $\pm$ 0.35}      & \sota{94.46 $\pm$ 0.37}      & \sota{65.37 $\pm$ 0.48}      & \sota{69.61 $\pm$ 0.72}      & \sota{94.01 $\pm$ 0.54}       & \sota{93.91 $\pm$ 0.61}      & \sota{33.28 $\pm$ 0.32}      & \sota{55.13 $\pm$ 0.24}      \\
                \bottomrule
            \end{tabular}}
        \begin{tablenotes}\footnotesize
            \item $^\dagger$ The best and runner-up results are marked as \textbf{bolded} and \underline{underlined}, respectively.
        \end{tablenotes}
    \end{threeparttable}
    \label{tab:nodeclas_main}
\end{table*}

\begin{table}[t]
    \caption{Quantitative results (\%) on the RCDD dataset. ``OOM'' indicates that the model runs out of memory on a V100 GPU with 32GB memory.}
    \begin{tabular}{lccc}
        \toprule & Macro-F1                     & Micro-F1                     & AP                           \\
        \midrule
        GCN      & 91.30 $\pm$ 0.38             & 98.29 $\pm$ 0.12             & \underline{90.85 $\pm$ 0.30} \\
        GAT      & 89.75 $\pm$ 0.21             & 98.02 $\pm$ 0.04             & 86.95 $\pm$ 0.49             \\
        LINKX    & OOM                          & OOM                          & OOM                          \\
        FAGCN    & 90.62 $\pm$ 0.32             & 98.09 $\pm$ 0.07             & 88.95 $\pm$ 0.49             \\
        \hline
        RGCN     & \underline{92.25 $\pm$ 0.34} & \underline{98.30 $\pm$ 0.07} & 90.73 $\pm$ 0.36             \\
        HAN      & 87.32 $\pm$ 0.34             & 97.46 $\pm$ 0.07             & 83.00 $\pm$ 0.36             \\
        RGAT     & 89.19 $\pm$ 1.20             & 97.93 $\pm$ 0.16             & 87.58 $\pm$ 0.45             \\
        HGT      & 91.04 $\pm$ 0.48             & 98.29 $\pm$ 0.08             & 88.65 $\pm$ 1.03             \\
        SHGN     & 88.12 $\pm$ 0.52             & 97.68 $\pm$ 0.10             & 83.89 $\pm$ 0.73             \\
        HINormer & OOM                          & OOM                          & OOM                          \\
        \midrule
        \ours    & \sota{92.92 $\pm$ 0.17}      & \sota{98.79 $\pm$ 0.21}      & \sota{92.87 $\pm$ 0.33}      \\
        \bottomrule
    \end{tabular}
    \label{tab:rcdd}
\end{table}

\section{Experiments}

In this section, we perform experimental evaluations to demonstrate the effectiveness of our proposed \ours framework.
Due to space limitations, we present detailed experimental settings and additional results in Appendix~\ref{sec:appendix-exp_setting} and \ref{sec:appendix-result}.

\subsection{Experimental Setups}
\nosection{Datasets}
We evaluate our proposed model on five real-world heterogeneous graphs, including four common benchmark datasets DBLP~\cite{lv2021we}, IMDB~\cite{lv2021we}, ACM~\cite{wang2019heterogeneous}, and MAG~\cite{wang2020microsoft}, as well as one industrial-scale dataset RCDD~\cite{liu2023datasets}.
The detailed statistics of all datasets are shown in Table~\ref{tab:dataset}. We provide further details for the datasets in Appendix~\ref{sec:appendix-datasets}.

\nosection{Baselines}
To comprehensively evaluate the proposed \ours against the state-of-the-art approaches, we consider six HGNN baselines, including RGCN~\cite{schlichtkrull2018modeling}, RGAT~\cite{ishiwatari2020relation}, HAN~\cite{wang2019heterogeneous}, HGT~\cite{hu2020heterogeneous}, SHGN~\cite{lv2021we}, and HINormer~\cite{mao2023hinormer}.
In addition, we include four homogeneous GNNs including GCN~\cite{kipf2016semi}, GAT~\cite{velivckovic2017graph}, as well as LINKX~\cite{zhu2020beyond} and FAGCN~\cite{bo2021beyond} which are specifically designed for handling heterophilic graphs. Further details for these baselines are provided in Appendix~\ref{sec:appendix-baselines}.
The hyperparameters of all the baselines were configured according to the experimental settings officially reported by the authors and were then carefully tuned in our experiments to achieve their best results.

\nosection{Evaluation Settings}
We evaluate the classification performance of \ours and baselines using Macro-F1 and Micro-F1 metrics on four benchmark datasets: DBLP, IMDB, ACM, and MAG. Additionally, for the RCDD dataset, we include average precision (AP) as an evaluation metric, as suggested in \cite{liu2023datasets}. All experiments are repeated five times, and we report the averaged results with standard deviations.

\subsection{Node Classification Performance}
\label{sec:ssnc}
\begin{figure}[t]
    \centering
    \includegraphics[width=0.45\linewidth]{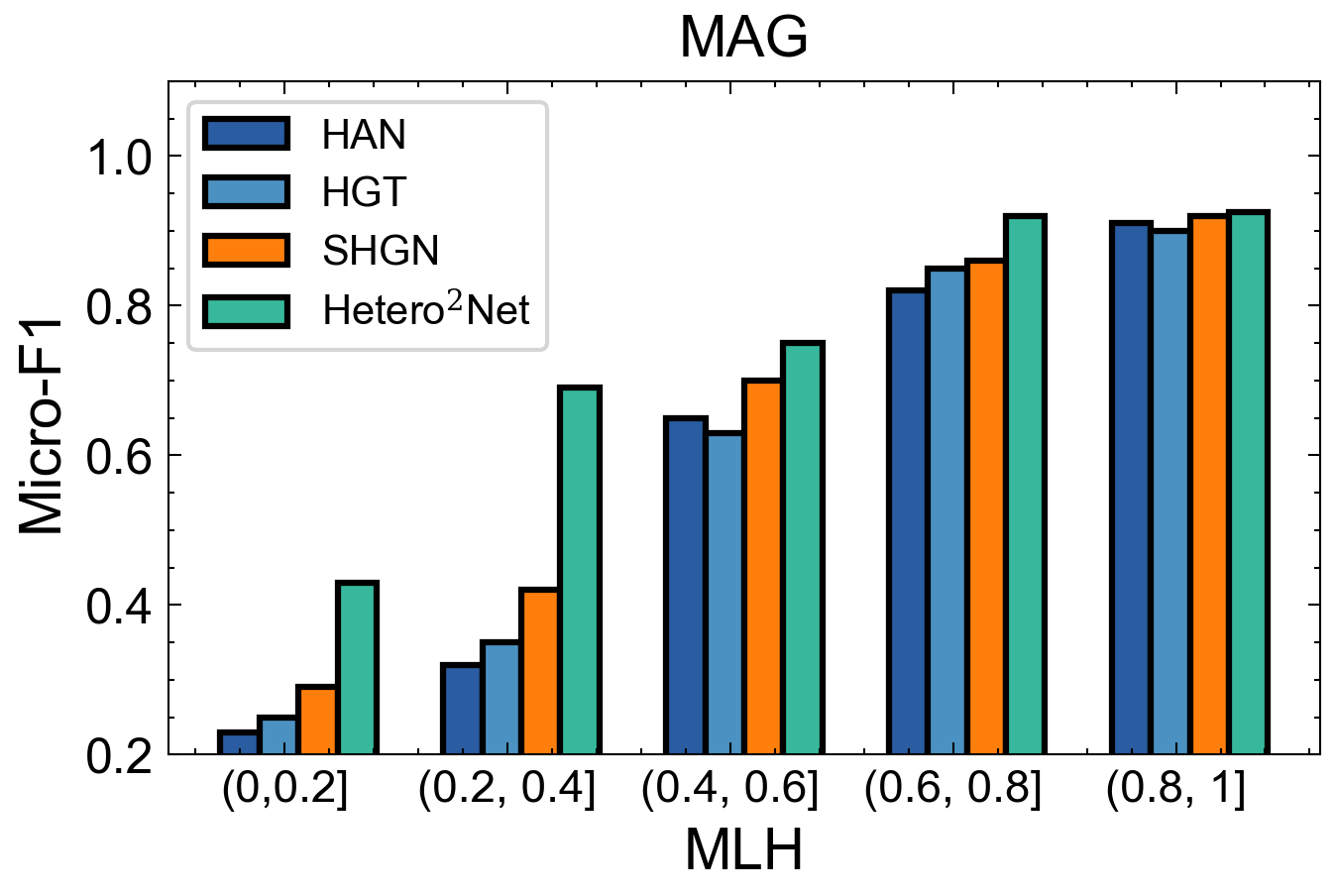}
    \includegraphics[width=0.45\linewidth]{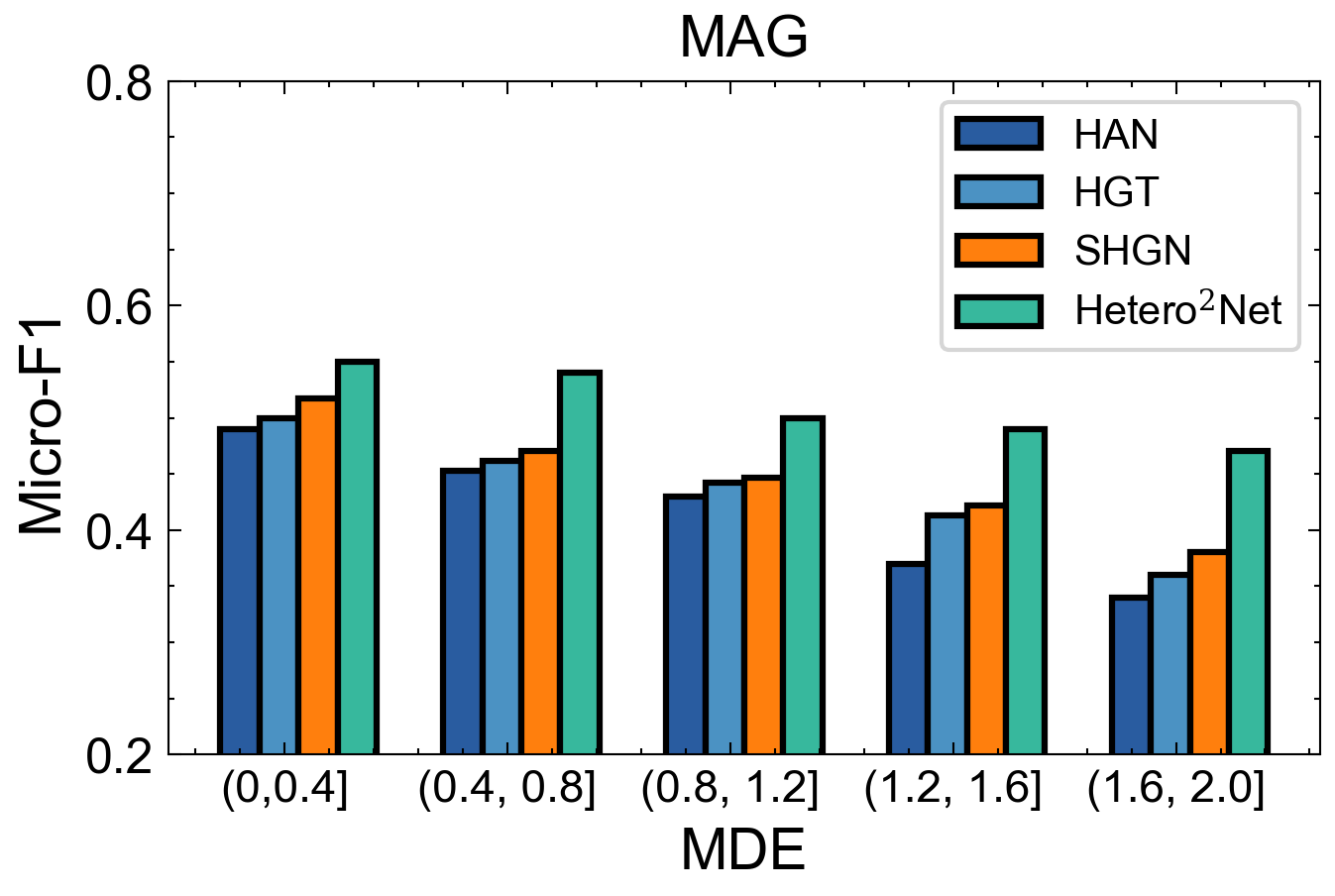}
    \vspace{-4mm}
    \caption{Performance comparison in terms of nodes with different MLH and MDE, respectively.}
    \label{fig:homophily_abla}
\end{figure}

We perform experiments on the node classification tasks using these five datasets. The overall performance comparisons on DBLP, IMDB, ACM, and MAG are presented in Table~\ref{tab:nodeclas_main}. Considering its industrial scale, we specifically highlight the effectiveness of our approach on the RCDD dataset, as indicated in Table~\ref{tab:rcdd}. Moreover, in Figure~\ref{fig:homophily_abla}, we show the performance comparison of our proposal and several representative HGNNs over nodes with varying levels of MLH and MDE. Based on these results, we can make the following observations.

\nosection{Comparison with Baselines} Table~\ref{tab:nodeclas_main} shows performance comparisons between \ours and the baselines on DBLP, IMDB, ACM, and MAG. For datasets with strong attribute-wise heterophily, DBLP and ACM, our proposal consistently outperforms the baselines across all metrics, which indicates the effectiveness of the disentanglement module in making full use of the node attributes to solve heterophily issues. Regarding datasets with strong label-wise heterophily, IMDB and MAG, \ours achieves a substantial improvement over all metrics, demonstrating the superiority of masked label prediction that can enhance message propagation by leveraging the label information.

Notably, the improvement is more significant in MAG compared to the other three datasets, as depicted in Table\ref{tab:nodeclas_main}. The observation ratios of target nodes in Table\ref{tab:dataset} present indicative clues for this phenomenon, with the MAG dataset exhibiting a substantially larger observation ratio of target nodes compared to other datasets. This finding aligns with the argument presented in~\cite{ShiHFZWS21}, which suggests that incorporating both node features and labels during training and inference stages can enhance model performance. This, in turn, serves as the primary explanation for the significant improvements achieved by \ours on MAG, highlighting the efficacy of the masked label prediction module.

\nosection{Performance comparison on RCDD}
To further validate the effectiveness of \ours, we conduct experiments on RCDD, an industrial-scale graph with 13M nodes and 15M edges. The RCDD dataset presents additional challenges including class imbalance, data noises, and multitude node and edge types. These challenges pose difficulties for existing HGNNs to achieve favorable performance. Table~\ref{tab:rcdd} presents a performance comparison between \ours and the baselines on the RCDD dataset. Notably, \ours consistently outperforms all other baselines by a substantial margin, providing compelling evidence for the effectiveness of our proposed method in addressing the complexities inherent in large-scale and heterogeneous graphs.

\nosection{Relationship between Metric and Performance}
According to the results shown in Figure~\ref{fig:homophily_abla}, it can be observed that \ours consistently achieves superior performance on nodes that exhibit strong homophily. This observation further confirms the strong correlations between our proposed metrics and the performance of HGNNs. Notably, \ours outperforms the baselines consistently across nodes with varying levels of metrics. This advantage is particularly evident on nodes with stronger heterophily, providing strong evidence that \ours can effectively address the heterophily issue compared to existing HGNNs.

Furthermore, an interesting finding is that the fundamental GNN model, GCN~\cite{kipf2016semi}, outperforms other heterophily-specific models in most cases, especially on datasets characterized by high label-wise heterophily and low attribute-wise heterophily, such as MAG and RCDD. This finding aligns with previous studies on homogeneous graphs~\cite{ma2021homophily}, indicating that the GCN model performs well on label-wise heterophilic graphs when labels have distinguishable distributions, implying low attribute-wise heterophily. Therefore, MLH and MDE can provide comprehensive and complementary insights into the impact of homophily on model performance.

\section{Conclusion}

In this paper, we investigate the heterophily property in heterogeneous graphs and introduce two metapath-based metrics, MLH and MDE, to quantify the heterophily of labels and attributes in heterogeneous graphs, respectively.
Our empirical results show that both metrics are informative in terms of correlating with HGNN performance.
Furthermore, we present \ours as a principled framework to learn over both homophlic and heterophilic graphs. \ours leverages masked metapath prediction and masked label prediction as two auxiliary tasks to learn disentangled and informative representations for downstream tasks.
In our extensive experimental evaluation, \ours demonstrates superior performance compared to strong baselines in the semi-supervised node classification task. Particularly, \ours advances new state-of-the-art performance on an industrial graph dataset RCDD with challenging scenarios.

\bibliographystyle{ACM-Reference-Format}
\bibliography{main}
\newpage
\appendix
\section{Algorithm}
\label{sec:appendix-algo}
The algorithm description of \ours is shown in Algorithm~\ref{algo}.

\begin{algorithm}
    \caption{Heterophily-aware HGNN~(\ours)}
    \label{algo}
    \begin{flushleft}
        \textbf{Input:} The input graph $\mathcal{G}=\left\{\mathcal{V}, \mathcal{E}, \mathcal{A}, \mathcal{R}, \phi, \psi\right\}$, mathpath set $\mathcal{P}^*=\left\{\mathcal{P}_1,\mathcal{P}_2,\ldots\right\}$, node attributes $\left\{{x}_u, \forall u \in \mathcal{V}\right\}$, the number of layers $L$
        \\
        \textbf{Output:} The prediction probability $\left\{\hat{y}_u, \forall u \in \mathcal{V}\right\}$
    \end{flushleft}
    \begin{algorithmic}[1]
        \State // Mask the input graph based on metapaths
        \State Mask $\mathcal{G}$ to generate $\mathcal{G}_\text{mask}$ and $\mathcal{G}_\text{vis}$;
        \For {$u \in \mathcal{V}$}
        \State // Masked label prediction strategy
        \State Generate enhanced features by Eq.~\eqref{eq:lp};
        \State // Separate by relations into multiple views
        \For {each view of relation $\psi(u,v)\in \mathcal{R}$}
        \State // Execute aggregation and combination for $u$
        \State Calculate $h_u^{(l)}(\psi(u,v))$ by Eq.~\eqref{eq:agg};
        \State // Fuse representations from all views
        \State Calculate $h_u^{(l)}(\phi(u))$ by Eq.~\eqref{eq:fuse};
        \EndFor
        \State // Decouple homophilic and heterophilic representations
        \State $\hat{h}_u(\psi(u,v)) = \hat{h}_u^\text{(homo)} + \hat{h}_u^\text{(hetero)}$;
        \State Calculate correlation loss $\mathcal{L}_\text{corr}$ by Eq.~\eqref{eq:corr};
        \State // Reconstruct $\mathcal{G}_\text{mask}$ by $\mathcal{G}_\text{vis}$
        \State Calculate reconstruction loss $\mathcal{L}_\text{rec}$ by Eq.~\eqref{eq:rec};
        \State Calculate classification loss by Eq.~\eqref{eq:cls};
        \State // Obtain overall objection function $\mathcal{J}$
        \State Calculate $\mathcal{J}$ by Eq.~\eqref{eq:obj};
        \EndFor
        \State $\hat{y}_u \leftarrow \textsc{Softmax}(h_u^{(L)}), \forall u \in \mathcal{V}$;
    \end{algorithmic}
\end{algorithm}

\begin{table}[h]
    \caption{Metapath-wise homophily metrics of datasets.}
    \label{tab:metapath}
    {\begin{tabular}{l|cccc}
            \toprule
            \textbf{Datasets}     & \textbf{Metapath $\mathcal{P}$} & $\mathcal{H}(\mathcal{G}_{\mathcal{P}})$ & $\mathscr{G}(\mathcal{G}_{\mathcal{P}})$ \\
            \midrule
            \multirow{1}{*}{DBLP} & \textit{APA}                    & 0.87                                     & 0.80                                     \\ \midrule
            \multirow{3}{*}{IMDB} & \textit{MDM}                    & 0.40                                     & 1.21                                     \\
                                  & \textit{MAM}                    & 0.17                                     & 0.44                                     \\
                                  & \textit{MKM}                    & 0.13                                     & 0.13                                     \\ \midrule
            \multirow{3}{*}{ACM}  & \textit{PAP}                    & 0.81                                     & 2.16                                     \\
                                  & \textit{PSP}                    & 0.64                                     & 0.04                                     \\
                                  & \textit{PTP}                    & 0.33                                     & 0.01                                     \\ \midrule
            \multirow{2}{*}{MAG}  & \textit{PFP}                    & 0.09                                     & 0.03                                     \\
                                  & \textit{PAP}                    & 0.34                                     & 0.05                                     \\ \midrule
            \multirow{2}{*}{RCDD} & \textit{IFI}                    & 0.66                                     & 0.14                                     \\
                                  & \textit{IBI}                    & 0.26                                     & 0.19                                     \\
            \bottomrule
        \end{tabular}
    }

\end{table}

\section{Detailed experimental settings}
\label{sec:appendix-exp_setting}
\subsection{Further Details of Datasets}
\label{sec:appendix-datasets}
In our experiments, we consider a total of five real-world heterogeneous graphs: three academic citation datasets, including DBLP~\cite{lv2021we}, ACM~\cite{wang2019heterogeneous}, and MAG~\cite{wang2020microsoft}; one movie-related dataset, IMDB~\cite{lv2021we}; and one commodity-related dataset, RCDD~\cite{liu2023datasets}. Table~\ref{tab:metapath} presents the \textit{length-2 metapaths} of these datasets, along with the metapath-wise homophily metrics. The overall homophily metrics for each dataset can be calculated by averaging the corresponding metapath-wise metrics, as shown in Eq.~\eqref{eq:metrics}.

Here are the details of the datasets:
\begin{itemize}
    \item \textbf{DBLP}\footnote{http://web.cs.ucla.edu/~yzsun/data/} is a bibliography website focusing on computer science. It contains multiple node types, including author~(A), paper~(P), and others. In our experiments, we concentrate on the classification of authors, and the corresponding \textit{length-2 metapath} is \textit{APA}.
    \item \textbf{IMDB}\footnote{https://www.kaggle.com/karrrimba/} is a website providing movie-related information. It includes four types of nodes: movie~(M), director~(D), actor~(A), and keyword~(K). In our experiments, we focus on the classification of movies, and the corresponding \textit{length-2 metapaths} are \{\textit{MDM}, \textit{MAM}, \textit{MKM}\}.
    \item \textbf{ACM}\footnote{http://dl.acm.org/} is a citation network that consists of four node types: paper~(P), author~(A), subject~(S), and term~(T). In our experiments, we focus on the classification of papers, and the corresponding \textit{length-2 metapaths} are \{\textit{PAP}, \textit{PSP}, \textit{PTP}\}.
    \item \textbf{MAG}\footnote{https://ogb.stanford.edu/docs/nodeprop/} is another academic network with multiple node types, including paper~(P), field\_of\_study~(F), author~(A), and others. In our experiments, we concentrate on the classification of papers, and the corresponding \textit{length-2 metapaths} are \{\textit{PFP}, \textit{PAP}\}.
    \item \textbf{RCDD}\footnote{https://zenodo.org/record/8103003} is a risk commodity detection dataset. Its node types consist of item~(I), nodes named `f'~(F), nodes named `b'~(B), and others. In our experiments, we focus on the classification of `f', and the corresponding \textit{length-2 metapaths} are \{\textit{PFP}, \textit{PAP}\}.
\end{itemize}

\begin{table*}[t]
    \caption{Ablation studies with \ours.}
    \begin{threeparttable}
        \resizebox{\linewidth}{!}
        {
            \begin{tabular}{lcccccccc}
                \toprule                                               & \multicolumn{2}{c}{ DBLP } & \multicolumn{2}{c}{ IMDB } & \multicolumn{2}{c}{ACM} & \multicolumn{2}{c}{ MAG }                                                                                                         \\
                \cmidrule{2-9}                                         & Macro-F1                   & Micro-F1                   & Macro-F1                & Micro-F1                  & Macro-F1                & Micro-F1                & Macro-F1                & Micro-F1                \\
                \midrule
                \ours                                                  & \sota{94.03 $\pm$ 0.35}    & \sota{94.46 $\pm$ 0.37}    & \sota{65.37 $\pm$ 0.48} & \sota{69.61 $\pm$ 0.72}   & \sota{94.01 $\pm$ 0.54} & \sota{93.91 $\pm$ 0.61} & \sota{33.28 $\pm$ 0.32} & \sota{55.13 $\pm$ 0.24} \\

                $-\mathcal{L}_\text{corr}$                             & 93.33 $\pm$ 0.42           & 93.94 $\pm$ 0.55           & 64.17$\pm$ 0.25         & 68.19 $\pm$ 0.t2          & 93.45 $\pm$ 0.50        & 93.38 $\pm$ 0.46        & 32.59 $\pm$ 0.10        & 53.10 $\pm$ 0.18        \\
                $-\mathcal{L}_\text{rec}$                              & 92.65 $\pm$ 0.65           & 93.09 $\pm$ 0.47           & 64.22 $\pm$ 0.32        & 68.33 $\pm$ 0.59          & 93.75 $\pm$ 0.44        & 93.62 $\pm$ 0.53        & 31.32 $\pm$ 0.16        & 52.11 $\pm$ 0.15        \\
                $-(\mathcal{L}_\text{corr} \& \mathcal{L}_\text{rec})$ & 92.94 $\pm$ 0.58           & 93.26 $\pm$ 0.62           & 63.88 $\pm$ 0.40        & 67.80 $\pm$ 0.74          & 93.49 $\pm$ 0.39        & 93.43 $\pm$ 0.71        & 31.19 $\pm$ 0.14        & 51.97 $\pm$ 0.19        \\
                $-\text{MaskLP}$$^\dagger$                             & -                          & -                          & 64.15 $\pm$ 0.25        & 68.41 $\pm$ 0.24          & 93.58 $\pm$ 0.35        & 93.44 $\pm$ 0.08        & 28.23 $\pm$ 0.21        & 49.04 $\pm$ 0.09        \\
                \bottomrule
            \end{tabular}

        }
        \begin{tablenotes}\footnotesize
            \item $^\dagger$MaskLP: masked label prediction task.
        \end{tablenotes}
    \end{threeparttable}
    \label{tab:ablation-study}
\end{table*}

\subsection{Further Details of Baselines}
\label{sec:appendix-baselines}
In this section, we provide more detailed information on the baseline methods. We categorize these methods into two main groups: homogeneous graph neural networks and heterogeneous graph neural networks.

\textit{(1) Homogeneous graph neural networks}
\begin{itemize}
    \item GCN~\cite{kipf2016semi} is a fundamental GNN that utilizes a localized first-order approximation of spectral graph convolutions to design a graph convolutional network.
    \item GAT~\cite{velivckovic2017graph} is a GNN that employs the attention mechanism to aggregate node features.
    \item LINKX~\cite{zhu2020beyond} is a GNN that decouples structure and feature transformation, making it simple and scalable.
    \item FAGCN~\cite{bo2021beyond} is a GNN with a self-gating mechanism, which can adaptively integrate different signals during message passing.
\end{itemize}

\textit{(2) Heterogeneous graph neural networks}
\begin{itemize}
    \item RGCN~\cite{schlichtkrull2018modeling} is a HGNN that introduces relation-specific transformations to separately aggregate neighbors based on relations.
    \item RGAT~\cite{ishiwatari2020relation} is a HGNN that incorporates relational position encodings to capture sequential information reflecting the relational graph structure.
    \item HAN~\cite{wang2019heterogeneous} is a HGNN that designs node-level and semantic-level attentions to aggregate neighborhood information along different metapaths.
    \item HGT~\cite{hu2020heterogeneous} is a HGNN that introduces an attention mechanism to maintain node- and edge-type dependent representations.
    \item SHGN~\cite{lv2021we} is a HGNN that leverages type information through learnable type embeddings. Additionally, this method improves performance by incorporating residual connections and applying an $l_2$-norm on the output.
    \item HINormer~\cite{mao2023hinormer}
          is a HGNN that capitalizes on a larger-range aggregation mechanism for node representation learning by a local structure encoder and a heterogeneous relation encoder.
\end{itemize}

\subsection{Hyperparameters Settings}
\label{sec:appendix-imple}
We set the embedding dimension of \ours to 128 for small-scale datasets (DBLP, IMDB, ACM) and 256 for large-scale datasets (MAG and RCDD). The hyperparameters $\alpha$ and $\beta$ are tuned from a set of values \{0, 0.1, 0.2, 0.3, 0.4, 0.5\}. The masking ratio on node labels, denoted as $p$, is tuned from a set of values \{0.5, 0.6, 0.7, 0.8, 0.9, 1.0\}. It is worth noting that we do not adopt masked label prediction in the DBLP dataset, which exhibits strong label homophily.

\section{Additional empirical results}
\label{sec:appendix-result}
\subsection{Ablation Study}

To evaluate the contribution of each component in \ours, we conduct extensive experiments of ablation study on four benchmark datasets by comparing with several degenerate variants: (1) $-\mathcal{L}_\text{corr}$: we remove the correlation loss and only use the construction loss in disentanglement module; (2) $-\mathcal{L}_\text{rec}$: we remove the construction loss and only consider correlation loss in disentanglement module; (3) $-(\mathcal{L}_\text{corr} \& \mathcal{L}_\text{rec})$: we only maintain the classification loss and remove other losses; (4) $-\text{MaskLP}$: we remove the masked label prediction module and preserve disentangled representation learning with masked metapath prediction.

Based on the results of the ablation studies presented in Table~\ref{tab:ablation-study}, several key observations can be made. Overall, both the learning tasks for disentanglement and masked label prediction consistently enhance the performance of \ours across all datasets. Specifically, the integrated utilization of the two learning tasks demonstrates superior performance in most cases within the disentanglement module. Furthermore, the exclusion of the masked metapath prediction technique leads to a notably larger performance degradation for the MAG dataset compared to the other datasets, thus corroborating the observation discussed in Section~\ref{sec:ssnc}.

\section{A discussion on the scale of Dirichlet energies}
\label{sec:discussion-energy}
In this section, we elaborate more on using Dirichlet energies as a measure of homophily. We will start from the scale of Eq. ~\eqref{eq:energy} on homogeneous graphs, extensions to metapath-induced graphs are straightforward. The definition is closely related to the renowned Laplacian quadratic form with respect to the normalized Laplacian \cite{maskey2023fractional} by
\begin{align}
    \mathscr{E}_\text{edge}(\mathcal{G}) = \frac{1}{2}\text{trace}\left(X^T L X\right),
\end{align}
where $L = I - D^{-1/2}A D^{-1/2}$ is the normalized graph Laplacian. The above definition suggests that if the scale of the feature matrix $X$ is properly controlled, then the scale of the energy will be well-controlled by the variational definition of eigenvalues:
\begin{align}
    \mathscr{E}_\text{edge}(\mathcal{G}) \le \frac{\left\| L\right\|_\text{op}}{2} \le 1\quad \text{if} \left\| X\right\|_2 \le 1.
\end{align}
Where we use $\left\|\cdot\right\|_\text{op}$ to denote the operator norm. In practice, the feature matrix might have a Frobenous norm greater than $1$, resulting in energies larger than $1$. Yet as illustrated in table \ref{tab:metapath}, the energy of actual graphs (or graph views) are well controlled by some moderate constants which alleviate the need to conduct normalization procedures that might make the graph learning procedure much harder due to the distortion of feature scale. \par
Next, we discuss node-level notions of Dirichlet energy which is written as:
\begin{align}
    \begin{aligned}
        \mathscr{E}_\text{node}(\mathcal{G}) & = \frac{1}{|\mathcal{V}|}\sum_{u \in \mathcal{V}} \mathscr{E}(u)                                                                                                                                                    \\
                                             & = \frac{1}{|\mathcal{V}|} \sum_{u \in \mathcal{V}} \frac{1}{4|\mathcal{N}_u|} \sum_{v \in \mathcal{N}_u}\left\|\frac{\mathrm{x}_u}{\sqrt{|\mathcal{N}_u|}}-\frac{\mathrm{x}_v}{\sqrt{|\mathcal{N}_v|}}\right\|_2^2.
    \end{aligned}
\end{align}
To obtain a scale upper bound on $\mathscr{E}(u)$, it suffices to use $1$ as a neighborhood size lower bound and apply Cauchy-Schwartz to arrive at $\mathscr{E}(u) \le \frac{M}{2}$, where $M$ is an $\ell_2$-norm upper bound on \emph{per-node} feature. This bound is much more realistic than controlling the Frobenius norm of the feature matrix, thereby justifying our choice of adopting node-level metrics. As we see from section \ref{sec:empiricals}, $M$ is well controlled in real-world datasets by some moderate constants, and once again we need no extra normalization procedures for our empirical evaluations.
\end{document}